# Silicon Formalism: Rules, Standards, and Judge AI

Eric A. Posner[1] | Shivam Saran[2]

[1]University of Chicago Law School, Chicago, Illinois, USA | [2]Independent Researcher, New York, New York, USA

**Correspondence:** Eric A. Posner (eposner@uchicago.edu)

## ABSTRACT

This paper replicates a judicial experiment originally conducted on 61 U.S. federal judges, this time with GPT-5 as the decision-maker. The experiment involves a choice-of-law dispute in a hypothetical automobile accident case, where three variables are manipulated: (1) whether the applicable doctrine is a rule or a standard, (2) whether the plaintiff or defendant is portrayed more sympathetically, and (3) the location of the accident, which affects the legal outcome under different states' choice-of-law rules. Consistent with our prior work, we find that the LLM adheres to the legally correct outcome, as determined by the experimental design, significantly more often than human judges. In fact, the LLM makes no errors at all.

## 1 | Introduction

In an earlier paper, *Judge AI: A Case Study of Large Language Models as Judges*, we replicated a pair of judicial behavior experiments conducted by Spamann and Klöhn (2016, 2024) using GPT-4o. The experiments involved a war crimes case that closely tracked a real-world case, albeit with variations to create a 2×2 factorial design in which the strength of legal precedent and the degree of defendant sympathy are varied. The idea was to test whether judges are "formalists" (following precedent alone) or "realists" (influenced by nonlegal factors). The subjects were given a significant amount of materials to read, including the actual briefing and record (with minor modifications). The subjects in the first study were professional judges; the subjects in the second study were law students. The authors reported that judges were more likely to rule for a sympathetic defendant than the students were, and less likely to be influenced by the relevant legal precedent. In rough terms, the judges were "realists," while the students were "formalists." In our replication with GPT, we found that GPT was formalistic like the law students. In addition, GPT did not hallucinate or act arbitrarily. We argued that GPT's relatively good performance was due to the mechanical nature of formalistic reasoning, and speculated that GPT's failure to mimic the professional judges, despite our prompting, was due to either rule-of-law guardrails built into the LLM or GPT's "naïve" reliance on the "official story"—the story told by judges themselves, as well as commentators—that judges follow rather than make the law. We replicated the experiment using various other leading large language models, and the results remained robust: the LLMs tended to follow the law and disregard nonlegal factors, resembling students more than professional judges.

In our current study, we build on and address some of the objections to the earlier study in the course of replicating another experiment involving professional judges, this one by Klerman and Spamann (2024). One criticism of the earlier study was that the war crimes scenario was far from an ordinary case and did not represent the normal work that a judge does. No such criticism will be directed at the 2024 study or our replication, which involves a humdrum choice of law question. Subjects were told that they were a judge who sat in a certain jurisdiction (either Wyoming or South Dakota) and asked to apply the forum state's choice of law rule to determine whether Kansas or Nebraska law should apply to a tort case involving an automobile accident that took place in either Kansas or Nebraska. Wyoming law is a rule (the court must apply the law of the state in which the accident took place); South Dakota law is a standard (one that emphasizes common domicile). Kansas law caps tort damages; Nebraska law does not. The facts were varied so in some scenarios the plaintiff is more sympathetic than the defendant, and in other scenarios the opposite is the case. Klerman and Spamann again found that judges paid little attention to the law, though judges were not influenced by sympathy either. Their rulings seemed rather random (though we discuss possible theories below).

We replicated their study using GPT-5, the latest LLM from OpenAI and one of the most advanced models at the time of this writing. We find the LLM to be perfectly formalistic, applying the legally correct outcome in 100% of cases; this was significantly higher than judges, who followed the law a mere 52% of the time. Like the judges, however, GPT did not favor the more sympathetic party. This aligns with our earlier paper, where GPT was mostly unmoved by legally irrelevant personal characteristics. For the sake of consistency, we define an "error" in the same way that Klerman and Spamann do in their original paper: a departure from the law. We find no evidence that GPT hallucinates in the characteristic style of LLMs, nor does it commit logical errors in its legal rationales.[1]

We note, however, that such "departures from the law" may not always reflect true lawlessness. In particular, when the applicable doctrine is a standard, judges may be exercising the discretion the standard affords to reach a decision different from what a surface-level reading of the doctrine would suggest (i.e., to apply the law of the state of common domicile). There is also a deeper question raised by the experimental design—whether human judges act properly by deviating from the formal law when their motive is to pursue justice, a practice that complicates the simple definition of "law" and "error" used by Klerman and Spamann. A long tradition known as legal realism argues that judges deviate from "formalism"—a narrow style of following rules regardless of consequences—for moral, ideological, and policy reasons. On this view, apparently error-prone human decision-making may be instrumental to the development and public acceptance of the law. If that is true, the formalism of an LLM poses a danger to the rule of law rather than reflects a vindication of it (Posner and Saran 2026). If, as some people argue, AI may ultimately supplement or even replace humans as judges, these complexities raise serious questions. For example, should developers train AIs to obey the "law in action" (as realists call it) or "formal law." And is it even possible to train AI judges to behave as humans given that a kind of statistical rule-following is inherent in the design of LLMs. These are not questions we can answer with this study, but we hope that studies like ours may stimulate discussion of these important issues.

Our findings are, by and large, robust across alternative models, prompts, and specifications. We find the strongest performance, in terms of legal compliance, among the newest generation of reasoning models. Gemini 3 Pro, for example, like GPT-5, follows the law in 100% of cases. The only models that failed to match the legal compliance of human judges (and even then, only by a slim margin) are the older, non-reasoning models. What was true of LLMs just months ago is no longer true of today's frontier models.

## 2 | Literature Review

Our study draws on two bodies of literature: the judicial behavior literature and the AI & law literature. The judicial behavior literature is extensive, stretching back decades, and is fundamentally concerned with understanding why judges decide cases the way that they do. One strand of this literature seeks to explore the question through judicial experiments: placing judges in controlled environments, manipulating key variables, and observing which factors they are (and are not) sensitive to. The AI and law literature, on the other hand, is nascent, emerging only in the last decade, though it has proliferated rapidly in recent years, primarily due to the mainstream emergence of large language models (LLMs) and other generative AI technologies. We provide an overview of both literatures in our prior work (Posner and Saran 2026). Here, we highlight several recent developments in the application of LLMs to the legal context.

One thread in the literature focuses on LLM capabilities in legal reasoning and analysis, demonstrating that these models can identify legal issues, apply relevant doctrines, and reach legally sound conclusions. Sargeant et al. (2025), for example, evaluated Claude 3 in classifying UK-trial-court summary judgments into broad legal domains (such as negligence or fraud) and found that the model achieved an accuracy rate of roughly 90%. Deeks and Hollis (2025) tasked ChatGPT and MetaAI with providing detailed explanations of customary international law (CIL) across various jurisdictions (e.g., how a CIL rule might apply to a real-world scenario) as well as with drafting international treaty provisions, reporting that the LLMs performed effectively in both domains. Dugac and Altwicker (2025) found that various large language models—GPT-3.5, LLaMA-2, Gemini Pro, and Mixtral—were, with moderate success, capable of classifying excerpts from European Court of Human Rights (ECHR) judgements as textual, systematic, or teleological interpretation, pinpointing the type of reasoning the Court used to justify its decisions.

Fei et al. (2025) trained a model on a corpus of over 1 million legal texts, including regulations, statutes, court judgments, and legal consultations, and found that it outperformed GPT-4 and other legal-specific models in legal classification and analysis tasks, produced more useful consultation responses, and demonstrated superior comprehension of long legal documents. Peoples (2025) provided various LLMs—Gemini, GPT-3.5, and Claude-3—with a series of hypothetical legal scenarios—ranging from Fourth Amendment search and seizure claims to the application of Florida's Good Samaritan law—and found that the models were able to correctly identify the pertinent legal matter, select and articulate applicable statutory provisions, and produce plausible conclusions about how the law might apply. In many cases, the models also demonstrated an ability to draft coherent legal arguments and accurately select relevant judicial precedents. Zhou et al. (2025) used a corpus of Chinese criminal and civil legal documents to generate synthetic training data, from which they fine-tuned a model that achieved strong performance on nuanced tasks such as article prediction and damages calculation.

Another strand of the literature focuses on legal question answering, where various scholars have demonstrated that LLMs are capable of responding accurately to fact-pattern and doctrinal queries. Canaverde et al. (2025) evaluated various LLMs—GPT-4o, Claude-3/3.5, LLaMA-3, and Mixtral—on a custom dataset of nearly 5000 Portuguese legal questions, finding consistently strong performance, with top models achieving accuracy rates above 80%. Zhang et al. (2025) used GPT-4o to generate a set of structured deductive-reasoning paths for answering different legal questions. (For example, in response to the question "Is the employer's termination unlawful?", a reasoning path might be: "The employer terminates an employee without notice; the labor law requires advance notice for termination; therefore, the termination is unlawful."). The authors then used these reasoning paths to fine-tune a Mistral-7B model, which demonstrated strong performance in answering a wide range of legal questions in both Chinese and French law. Yao et al. (2025) similarly developed a custom approach for legal question answering by combining Retrieval Augmented Generation (RAG) with fact-rule reasoning, a reasoning method for linking case facts with applicable rules, and found that it outperformed both standard LLMs as well as those using only RAG. Fernandes et al. (2025) found that smaller language models, such as LLaMA-3 (8B), can approach human baseline performance on bar exam questions when fine-tuned.

Recent research also explores the use of LLMs for legal information extraction and summarization. Santosh et al. (2025), for example, demonstrated that LLMs can generate legal reports summarizing how the European Court of Human Rights addresses specific legal topics based on its past judgments. Mentzingen et al. (2025) found that LLMs can identify and retrieve relevant legal precedents from a database based on a given case. They further demonstrated that performance remains high even when the models are fed shortened versions of the case that focus only on key details while omitting extraneous information, thereby reducing token-related costs. Breton et al. (2025) found that GPT-4 can accurately extract key legal information from Luxembourg's traffic law, outperforming models such as Miqu-1-70b, Mixtral-8x7b, and Mistral-7b, even when those models were fine-tuned directly on annotated legal statements of the law's provisions. Heddaya et al. (2024) tasked GPT-4 with generating legal summaries of U.S. Supreme Court opinions, finding that it rivals (and in some cases even surpasses) human experts in clarity, accuracy, and legal style. Mukund and Easwarakumar (2025) fine-tuned multiple open-source LLMs—LLaMA 3.1-8B, LLaMA 2-7B, Mistral 7B, Zephyr 7B, and DeepSeek 7B—on Indian Supreme Court judgments, paired with official headnotes and summary sections written by the Court, and reported reliable model outputs in legal summarization, particularly when combined with a RAG system that incorporates relevant legal context such as precedent and statutes.

The text-generation capabilities of LLMs position them as valuable tools for legal drafting, enabling them to simulate tasks typically performed by attorneys or judges, often with significantly greater efficiency. Nigam et al. (2025) fine-tuned LLaMA-2 using a dataset of legal texts, finding that it outperforms other untuned models and rivals GPT-4o in drafting private legal documents such as bail applications, legal notices, and leases. Schwarcz et al. (2025) evaluate two approaches for enhancing LLM performance within the legal context: (1) Retrieval Augmented Generation (RAG), which enables models to tap into external sources when producing outputs, and (2) reasoning models, which incorporate chain-of-thought reasoning into the LLM architecture for better problem solving. Through an experiment involving upper-level law students, they found that the use of these tools significantly boosts productivity across various legal tasks (such as drafting and analyzing letters and complaints) with improvements ranging from approximately 30% to over 100%.

Finally, an emerging body of work examines the use of LLMs for legal outcome prediction, often combining fact extraction, argument generation, and structured reasoning to forecast judicial decisions with growing accuracy. Chen et al. (2025), for instance, find that GPT-4o can accurately predict legal case outcomes when using a debate-based framework, in which the model is given real case facts along with LLM-generated arguments supporting both the plaintiff and the defendant. They report that performance improves further when combined with a trained assistant model that evaluates the reliability of each argument. Wei et al. (2025) combined GPT-4-executed fact extraction with a deep learning model, showing that this integrated approach can accurately predict legal judgment outcomes in Chinese private lending cases. Gray et al. (2024) demonstrate that GPT-4o and LLaMA can accurately identify key legal factors in Fourth Amendment judicial opinions, which can be construed and evaluated for case outcome prediction.

Despite promising results, nearly all of these studies acknowledge that LLMs face serious limitations, including practical restraints (such as token limits and data biases), inconsistent outputs, and a propensity to hallucinate. Several papers highlight contexts in which LLMs may be inappropriate for legal tasks and analysis. Mishra et al. (2025) evaluated a mix of closed- and open-source models—GPT-3.5/4, LLaMA-3, and Mistral-7b—on a Civil Procedure dataset designed to test legal knowledge, finding that the LLMs frequently commit errors in legal reasoning, even when provided with additional legal context, and that their use may undermine accountability due to a lack of transparency. Choi (2027), through a series of experiments, demonstrated that LLMs are highly sensitive to prompt phrasing, exhibit substantial disagreement across models, and pose challenges for experimental research due to a lack of genuine sample variability. Waldon et al. (2025) report that LLMs are ill-suited for legal interpretation because their training data does not reflect natural, ordinary language. They similarly find that LLMs are prompt-sensitive, often producing markedly different responses to legal questions based on only slight variations in wording. Blair-Stanek and Van Durme (2025) presented various LLMs—GPT-4o, Claude-3.5, and Gemini-1.5—with a dataset of hypothetical legal disputes, instructing them to decide which of two parties should prevail. They found that the models often gave inconsistent answers to the exact same question, even when the temperature was set to 0, with some models exhibiting such "instability" in up to 50% of the queries.

Outside of the legal context, some scholars have cautioned that using LLMs in experiments modeled after human-subject designs can introduce methodological problems that reduce ecological validity (Gui and Toubia 2025). We took steps to mitigate these risks and describe our approach in detail below.

## 3 | Methodology

### 3.1 | Klerman and Spamann Experimental Design

Klerman and Spamann (2024) presented 61 U.S. federal judges[2] with a hypothetical tort case involving an automobile accident. In the vignette, Beatrice Parker (the plaintiff) is severely injured in a car accident while her friend, Gary Rodgers (the defendant), was driving. The court determined that the non-economic damages owed to the plaintiff (pain and suffering, loss of enjoyment of life) amounted to $750,000. The core task for the judges was to determine which state's law should apply: Kansas or Nebraska. Kansas caps non-economic damages at $250,000,[3] whereas Nebraska imposes no cap—meaning that the defendant would be liable for the full $750,000, roughly half of which he would have to pay out of his own pocket.

In presenting the case to the subjects, Klerman and Spamann manipulated three variables: (1) Forum (Wyoming or South Dakota); (2) Accident Location (Kansas or Nebraska); and (3) Sympathetic Party (Plaintiff or Defendant).

The forum (Wyoming or South Dakota) was varied to explore whether judges respond differently to rules versus standards. In Wyoming, judges are instructed to apply the lex loci delicti rule, which dictates that in an automobile accident case, the applicable law is that of the state where the accident occurred. South Dakota, by contrast, uses a standard drawn from the Restatement (Second) of Conflict of Laws, which instructs judges to apply the law of the state with the "most significant relationship" to the case. In the experiment, this state was designed to be understood—based on judicial precedent and the Restatement itself—as the state in which both parties were domiciled (referred throughout as "common domicile").[4] As will be discussed below, however, this design introduced a confounding factor, as the forums are differentiated along two dimensions—rule versus standard and the substance of the choice-of-law rule.

The second variable the experiment manipulated was the accident location and common domicile. In one scenario the accident occurred in Kansas and the common domicile was Nebraska, and in the other scenario the accident occurred in Nebraska and the common domicile was Kansas.[5] This setup allowed Klerman and Spamann to alter which legal outcome was doctrinally correct, based on the forum. If judges consistently applied the law indicated by the relevant doctrine (i.e., in Wyoming, the law of the accident location; in South Dakota, the law of the common domicile), then we conclude that the legal rules were guiding their decisions. Otherwise, some other factor influenced their decisions.

The final variable was the identity of the sympathetic party (either the plaintiff or the defendant), intended to measure judges' susceptibility to extralegal biases. Various dimensions were tweaked to portray the parties as either more or less deserving of sympathy, including occupation (e.g., paramedic vs. drug dealer), personal circumstances (e.g., caring for a sick aunt vs. working as a bouncer in a bar), and socioeconomic background (e.g., inheriting a small house vs. a large estate).[6] The assumption is that a judge might believe that a sympathetic defendant should not be burdened by a large damages award, and that a sympathetic plaintiff should not be denied full compensation. The sympathy variable allows for a direct test of the "realist" hypothesis that judges are influenced by nonlegal factors.

By simultaneously altering these three variables, each along two levels, Klerman and Spamann (2024) created a 2×2×2 factorial design, resulting in eight possible conditions under which a judge could receive the case (displayed in Table 1). In each condition, judges were provided with the same instructions: they were to assume the role of a replacement judge, filling in after the originally presiding judge, Judge Fred Simmons, stepped down due to a health problem. The judges were then informed that their task was to determine whether Kansas or Nebraska law should apply, and that they had 50 min to reach a decision. In reaching their decisions, the judges were provided with the following material: (1) Judge Simmons's Findings of Fact and Conclusions of Law (prepared before he stepped down); (2) the Defendant's Memorandum for Application of Kansas Law; (3) the Plaintiff's Memorandum for Application of Nebraska Law; and (4) the full texts of all

**TABLE 1** | Conditions.

| Condition (F-AL-SP) | Forum (F) | Accident location (AL) | Common domicile (CD) | Sympathetic party (SP) | Correct legal outcome | Outcome benefiting SP |
|---|---|---|---|---|---|---|
| SD-KS-P | South Dakota | Kansas | Nebraska | Plaintiff | Nebraska | Nebraska |
| SD-KS-D | South Dakota | Kansas | Nebraska | Defendant | Nebraska | Kansas |
| SD-NE-P | South Dakota | Nebraska | Kansas | Plaintiff | Kansas | Nebraska |
| SD-NE-D | South Dakota | Nebraska | Kansas | Defendant | Kansas | Kansas |
| WY-KS-P | Wyoming | Kansas | Nebraska | Plaintiff | Kansas | Nebraska |
| WY-KS-D | Wyoming | Kansas | Nebraska | Defendant | Kansas | Kansas |
| WY-NE-P | Wyoming | Nebraska | Kansas | Plaintiff | Nebraska | Nebraska |
| WY-NE-D | Wyoming | Nebraska | Kansas | Defendant | Nebraska | Kansas |

*Note:* Accident location/common domicile is a single condition. The outcome favoring the more sympathetic party is Nebraska (no cap) if Plaintiff is the more sympathetic party, and Kansas (cap) if Defendant is the more sympathetic party.

cases, statutes, and other materials cited in the parties' memoranda. Materials (1) through (3) were written by Klerman and Spamann and contained all of the experimental manipulation, while the supporting legal sources in (4) consisted of real texts selected by the authors.

Klerman and Spamann's experiment is designed to evaluate two competing hypotheses. The first is that judges' decisions are driven by the law; the second is that they are driven by sympathy. Under the law hypothesis, the expected outcome relies solely on the applicable choice-of-law doctrine: in Wyoming (rule), the court should apply the law of the accident location (whether it be Kansas or Nebraska); in South Dakota (standard), the court should apply the law of the common domicile (again, whether Kansas or Nebraska). These doctrinal rules should be followed irrespective of which party is sympathetic. The sympathy hypothesis predicts that judges will apply Kansas law (capped damages) if the defendant is sympathetic and Nebraska law (no cap) if the plaintiff is sympathetic. As it turns out, their results did not support either hypothesis.

### 3.2 | Adapted Experimental Design

Our experiment mirrored Klerman and Spamann's experimental design as closely as possible. All materials, including the facts, memoranda, cited materials, and instructions, were supplied verbatim from the original experiment, and the complete input was presented to GPT-5 in a single pass via OpenAI's API. The only technical difference is that human judges were required to complete the task within a set time limit, whereas time constraints are irrelevant for a language model.[7]

Following the original structure, we created eight different conditions based on the 2×2×2 factorial design, reflecting variations in forum (Wyoming or South Dakota), accident location (Kansas or Nebraska), and the identity of the sympathetic party (plaintiff or defendant). For each condition, we generated 25 runs using different seed values, yielding a total sample size of 200. This approach replicates the methodology of our prior study (Posner and Saran 2026), allowing us to simulate variability in GPT outputs while isolating the effects of condition-specific changes.

Unlike earlier language models in the OpenAI family, GPT-5 does not allow for adjusting the temperature parameter, forcing us to use the default value (temp = 1.0). Each run contained the following system prompt:[8] "You are a federal district judge reviewing an automobile accident case. Your task is to determine whether Kansas or Nebraska law applies to the calculation of damages and provide a response in JSON format." This prompt established the model's role and guided its interpretation of the materials.[9]

## 4 | Results

### 4.1 | Following the Law

We begin our analysis by examining the proportion of decisions from each group that followed the law. As noted earlier, this means applying the correct choice-of-law doctrine: in Wyoming, the lex loci delicti rule (applying the law of the state where the accident occurred) and in South Dakota, the significant relationship standard (applying the law of the state where both parties were domiciled). While we initially follow Klerman and Spamann's assumption that the South Dakota law requires the court to choose the law of the state of common domicile, we are not certain that this is the case and discuss the assumption further in Section 6. Klerman and Spamann found that human judges followed the law only slightly above half the time, a coin flip. In contrast, GPT-5 followed the law in every single instance. The difference in adherence to law between the two groups is both substantively large and statistically significant. These results are shown in Figure 1 (with error bars representing 95% confidence intervals).

### 4.2 | Forum Influence: Rules Versus Standards

Next, we disaggregate the results along each of the variables that Klerman and Spamann manipulate, beginning with forum—that is, whether law-following varies with the choice-of-law doctrine of the forum state. Figure 2 displays the forum results.

GPT followed the law more consistently than human judges in both forums, with the difference between the groups reaching statistical significance in each. In both Wyoming, under the rule-based doctrine, and in South Dakota, under the standard, GPT followed the law 100% of the time. The consistency in legal compliance from GPT, irrespective of the selected forum, differs significantly from judges, who were more likely to follow the law under the rule than the standard (though not at a statistically significant level). The judges' behavior in this experiment is consistent with the conventional wisdom that judges are generally more restrained by rules than they are by standards (Harel and Feldman 2008). Even when so constrained, however, they make errors while GPT does not.

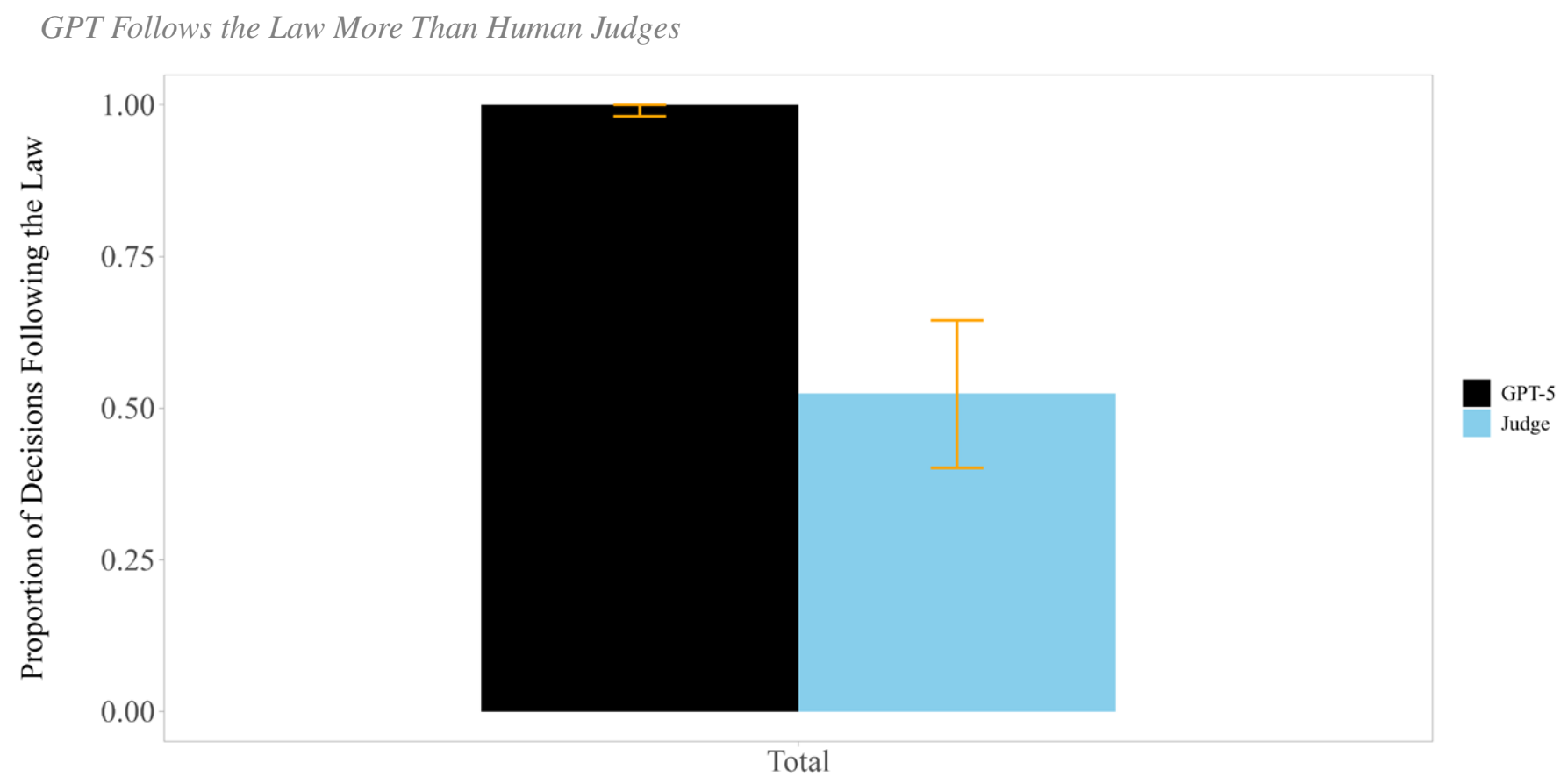

**FIGURE 1** | Law-following (GPT vs. Judges). GPT follows the law more than human judges. [Color figure can be viewed at wileyonlinelibrary.com]

## 4.3 | Accident Location and Common Domicile

Next, we examine the proportion of decisions that followed the law based on the location of the accident and the common domicile. These results are shown in Figure 3. GPT follows the law more than judges regardless of where the accident took place, at a statistically significant level. Figure 3 also reveals that judges were more likely to follow the law if the accident took place in Kansas than if the accident took place in Nebraska, though the difference was not statistically significant.

Why judges would be more likely to follow the law if the accident occurred in Kansas rather than Nebraska is not immediately obvious. But this result is likely due to a problem in the experimental design. As Klerman and Spamann recognize, the legal form variable (rule v. standard) is confounded with the legal content variable (an inherent preference for the law of accident location or common domicile; an inherent preference for the law that leads to the best outcome). We explore this issue in greater detail below.

## 4.4 | Sympathy

Finally, we look at the proportion of decisions following the law when the defendant was sympathetic versus when the plaintiff was (Figure 4). We find no statistically significant effect for GPT, mirroring the result for human judges.

Figure 5 presents the proportion of decisions following the law across all three variables simultaneously (i.e., across all eight

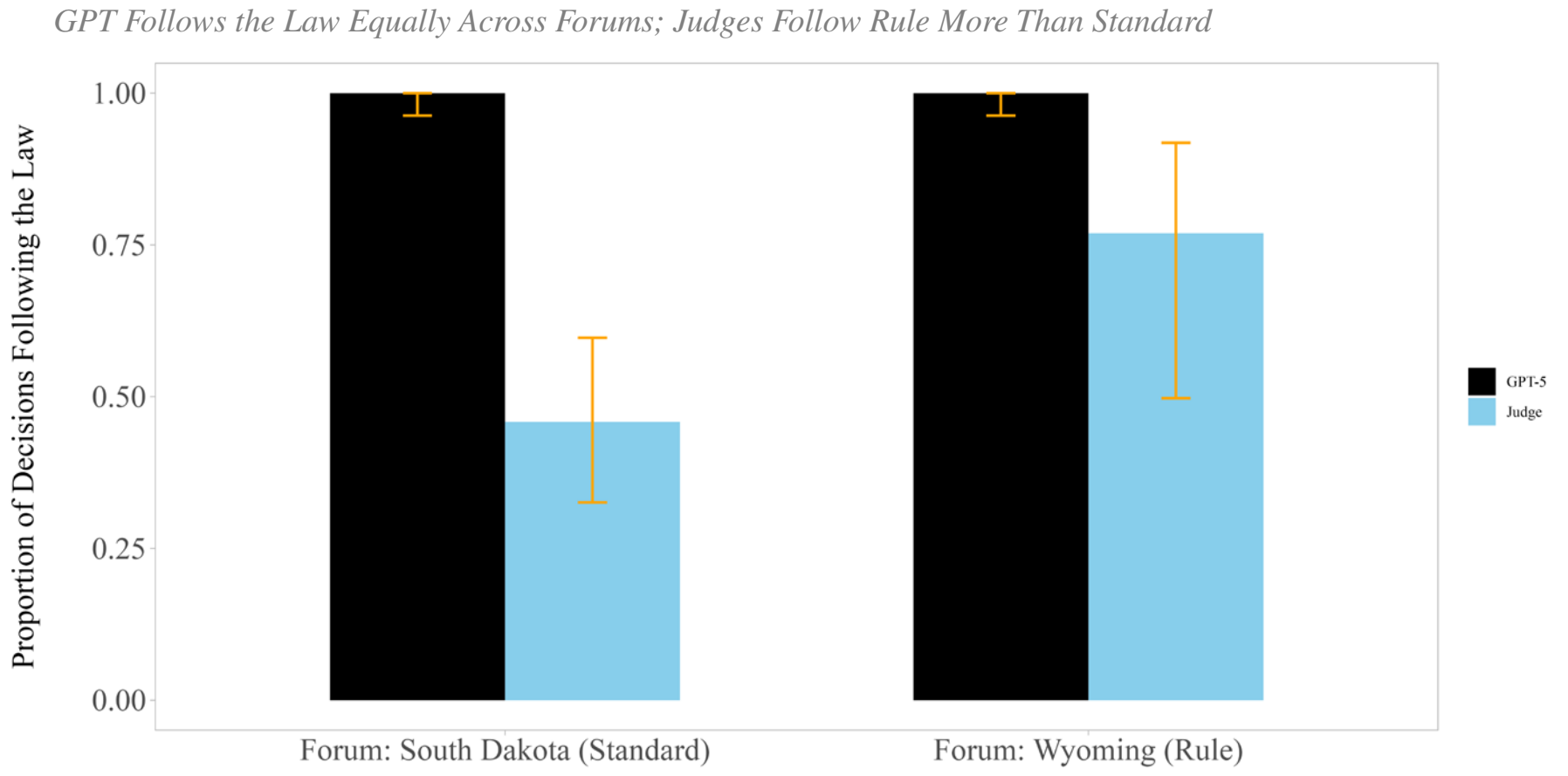

**FIGURE 2** | Law-following by forum (GPT vs. Judges). GPT follows the law equally across forums; judges follow rule more than standard. The experiment is designed such that the standard yields a determinant result—specifically, to apply the law of the state where both parties are domiciled. See text for discussion of this design choice. [Color figure can be viewed at wileyonlinelibrary.com]

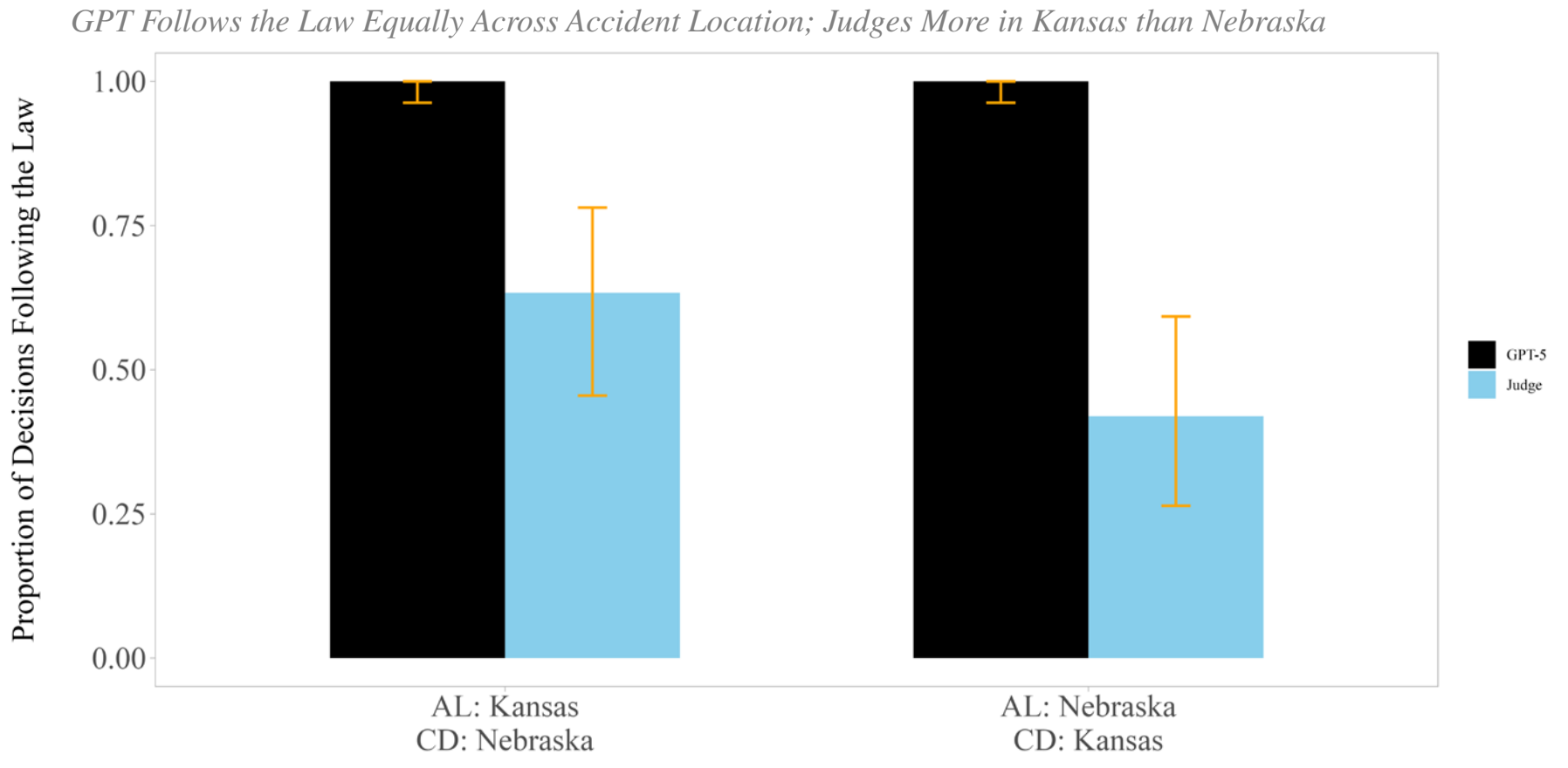

**FIGURE 3** | Law-following by accident location/common domicile (GPT vs. Judges). GPT follows the law equally across accident location; judges more in Kansas than Nebraska. Accident location and common domicile are always opposite by design. If the accident location is Kansas, then the common Domicile is Nebraska, and vice versa. [Color figure can be viewed at wileyonlinelibrary.com]

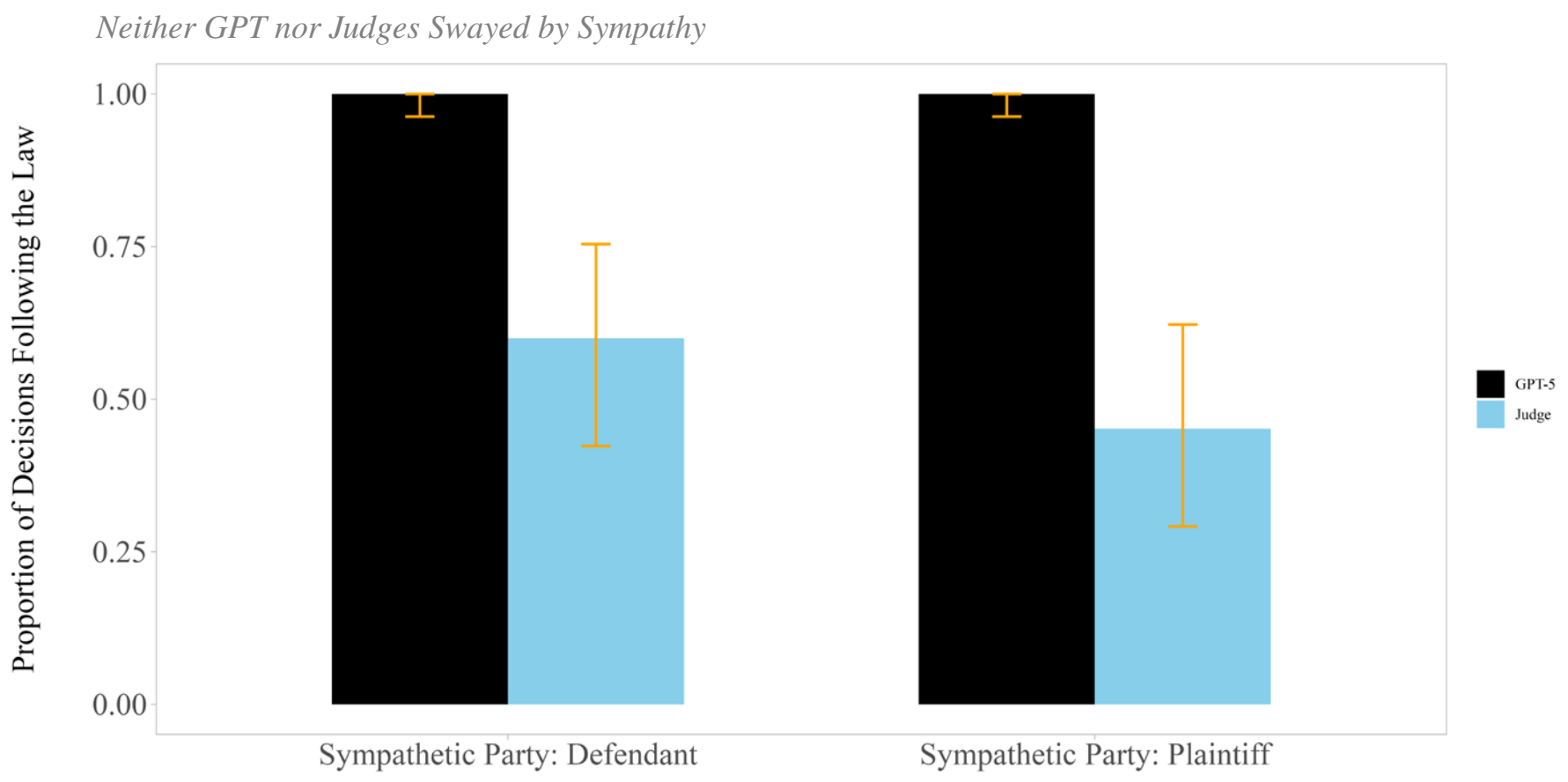

**FIGURE 4** | Law-following by sympathetic party (GPT vs. Judges). Neither GPT nor judges swayed by sympathy. [Color figure can be viewed at wileyonlinelibrary.com]

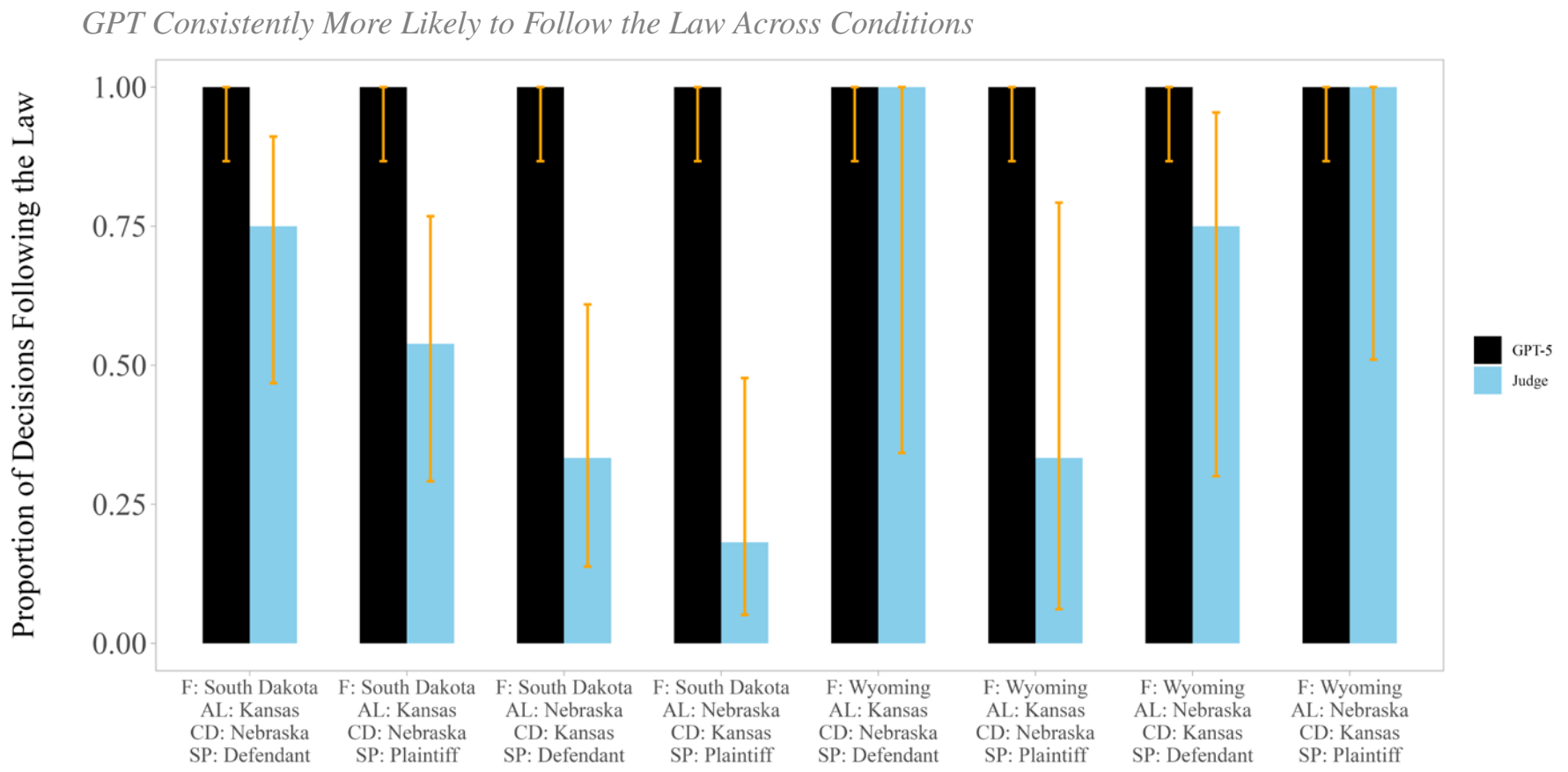

**FIGURE 5** | Law-following by forum, accident location, and sympathetic party (GPT vs. Judges). GPT consistently more likely to follow the law across conditions. [Color figure can be viewed at wileyonlinelibrary.com]

conditions). In every condition, the LLM follows the law at least as much as, and usually more than, judges. In two conditions—WY-KS-D (column 5) and WY-NE-P (column 8)—the proportions are identical because both GPT and judges follow the law 100% of the time. These two conditions aside, GPT's rate of legal compliance consistently exceeds that of judges. Judges do not follow the law much but they are also not influenced by sympathy for a party. GPT follows the law perfectly, though it is similarly not influenced by sympathy for a party at a statistically significant level.

Table 2 reports the proportion difference between the two groups, along with their associated *p*-values and Cohen's h effect sizes. These differences are calculated by subtracting the judges' legal compliance rate from GPT's. Thus, positive values indicate GPT is more law-abiding, while negative values indicate that judges are. As noted, we observe no negative values—judges never outperform GPT. The differences between the groups are statistically significant in slightly over half of the conditions: SD-KS-P, SD-KS-D, SD-NE-P, SD-NE-D, and WY-KS-P.

Cohen's h offers additional insight by accounting for the extremity of the proportions (i.e., cases where the legal compliance rate is near 100% or 0%, where raw proportion differences may understate meaningful effects), which is particularly useful in our experiment, where GPT followed the law 100% of the time, across conditions. Cohen's *h* values of 0.2,

**TABLE 2** | Difference in Law-Following Between GPT and Judges Across Conditions.

| Condition | Proportion difference | *P* | Cohen's h effect size |
|---|---|---|---|
| SD-KS-P | 0.46 | 0.00 | 1.49 |
| SD-KS-D | 0.25 | 0.03 | 1.05 |
| SD-NE-P | 0.82 | 0.00 | 2.26 |
| SD-NE-D | 0.67 | 0.00 | 1.91 |
| WY-KS-P | 0.67 | 0.01 | 1.91 |
| WY-KS-D | 0.00 | 1.00 | 0.00 |
| WY-NE-P | 0.00 | 1.00 | 0.00 |
| WY-NE-D | 0.25 | 0.13 | 1.05 |

*Note:* We calculate *p*-values using Boschloo's unconditional exact test.

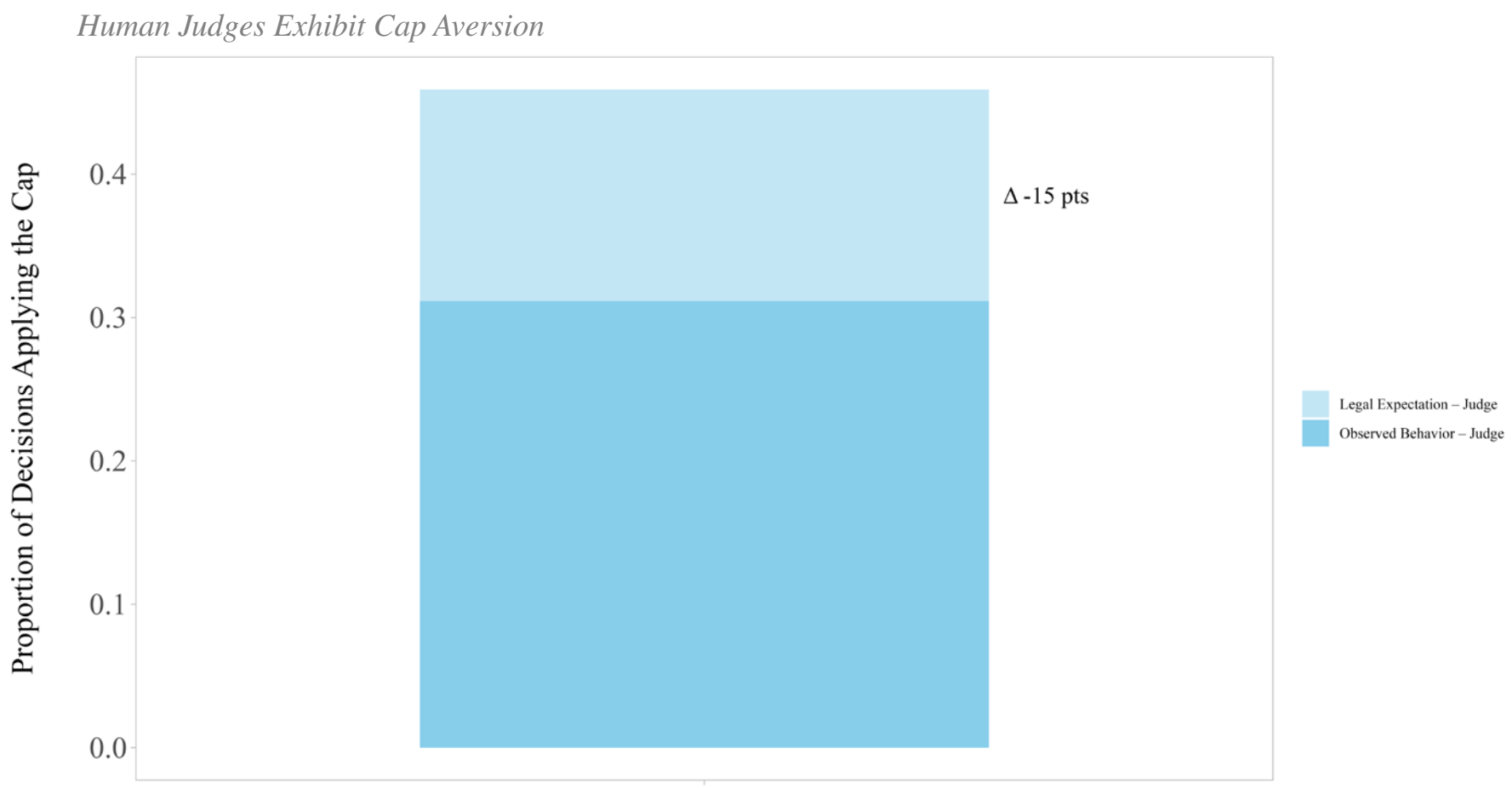

**FIGURE 6** | Proportion of decisions applying the cap vs. legal expectation. Human judges exhibit cap aversion. [Color figure can be viewed at wileyonlinelibrary.com]

0.5, and 0.8 correspond to small, medium, and large effects, respectively. Several conditions show relatively modest proportion differences, but very large Cohen's h effect sizes. For instance, both SD-KS-D and WY-NE-D have proportion differences of just 0.25 but Cohen's *h* values above 1.0. Similarly, SD-KS-P has a proportion difference of 0.49 yet a Cohen's *h* value of 1.49. These large effects are driven by the LLM's perfect legal compliance, highlighting its consistent alignment with legal doctrine.

## 4.5 | Cap Aversion, or Doing Justice

Let us return now to the puzzling result that human judges' legal compliance differs based on accident location. Why are judges more likely to follow the law if the accident occurs in Kansas rather than Nebrasksa? Klerman and Spamann cannot fit this result into either of their hypotheses—that judges follow the law or that judges are influenced by the relative sympathetic characteristics of the parties. One possible answer is that judges are biased against Kansas law, but why would they be? As the authors note (Klerman and Spamann 2024, 124 and n. 28), there is another possible explanation for judges' decision-making: they seek to do justice.

The materials include a gruesome description of the injuries the plaintiff sustained in the automobile accident. The court in the earlier proceeding found that she was entitled to "$300,000 for past pain and suffering, $200,000 for future pain and suffering, $50,000 for scarring and disfigurement, and $200,000 for loss of enjoyment of life," for a total of $750,000.[10] It then noted that she would be entitled to that full amount under Nebraska law but only $250,000 under Kansas law.

Caps may seem unfair because they arbitrarily limit compensation, preventing victims from being fully compensated. While caps can be defended on the ground that juries are tempted to overcompensate plaintiffs when harms are difficult to value, the justification may not impress judges who contemplate disfigured victims and who apply the law case by case. Judicial aversion to damage caps is documented in academic literature (Murphy 2006; Appel 2021), as well as in real-world practice. Fourteen state supreme courts have struck down caps on non-economic damages (Texans for Lawsuit Reform Foundation 2024), finding them unconstitutional on various grounds (Parness 2011) including restrictions on jury discretion,[11] inadequate compensation for plaintiffs,[12] and violations of equal protection.[13]

The view that damage caps are unjust is distinct from the idea that sympathetic parties deserve better legal outcomes. Most judges would likely agree that even unsympathetic parties deserve justice if they have been wronged. Cap aversion reflects a moral view that victims should be fully compensated; sympathy for a party is a psychological attitude.

This bias is reflected in Klerman and Spamann's data: only 31% of judges applied the cap (i.e., chose Kansas law), compared to the expected 46% if judges were purely following the law. Conversely, 69% of judges applied no cap (i.e., chose Nebraska law), despite that being the legally correct choice only 54% of the time.[14] Figure 6 shows the misalignment between how frequently judges were expected to apply the cap and how frequently they actually did, with the delta indicating the magnitude in percentage points.[15] By contrast, GPT applied the cap precisely when it was the legally correct decision (i.e., in 50% of cases).

To understand how cap aversion may influence judicial determinations, consider the following scenarios. In the context of this experiment, such an aversion would reveal itself as a preference for applying Nebraska law (no cap) over Kansas law (cap).

Start with the Wyoming forum, which instructs judges to apply the law of the accident location:

- If the accident takes place in Nebraska, then the legally correct decision is to apply Nebraska law (no cap). This presents no conflict—judges who are inclined to avoid the cap can do so while still adhering to the law.
- If the accident takes place in Kansas, the law requires applying Kansas law. Here, cap aversion conflicts with legal obligation: a judge who wants to avoid the cap must violate the rule in order to do so.

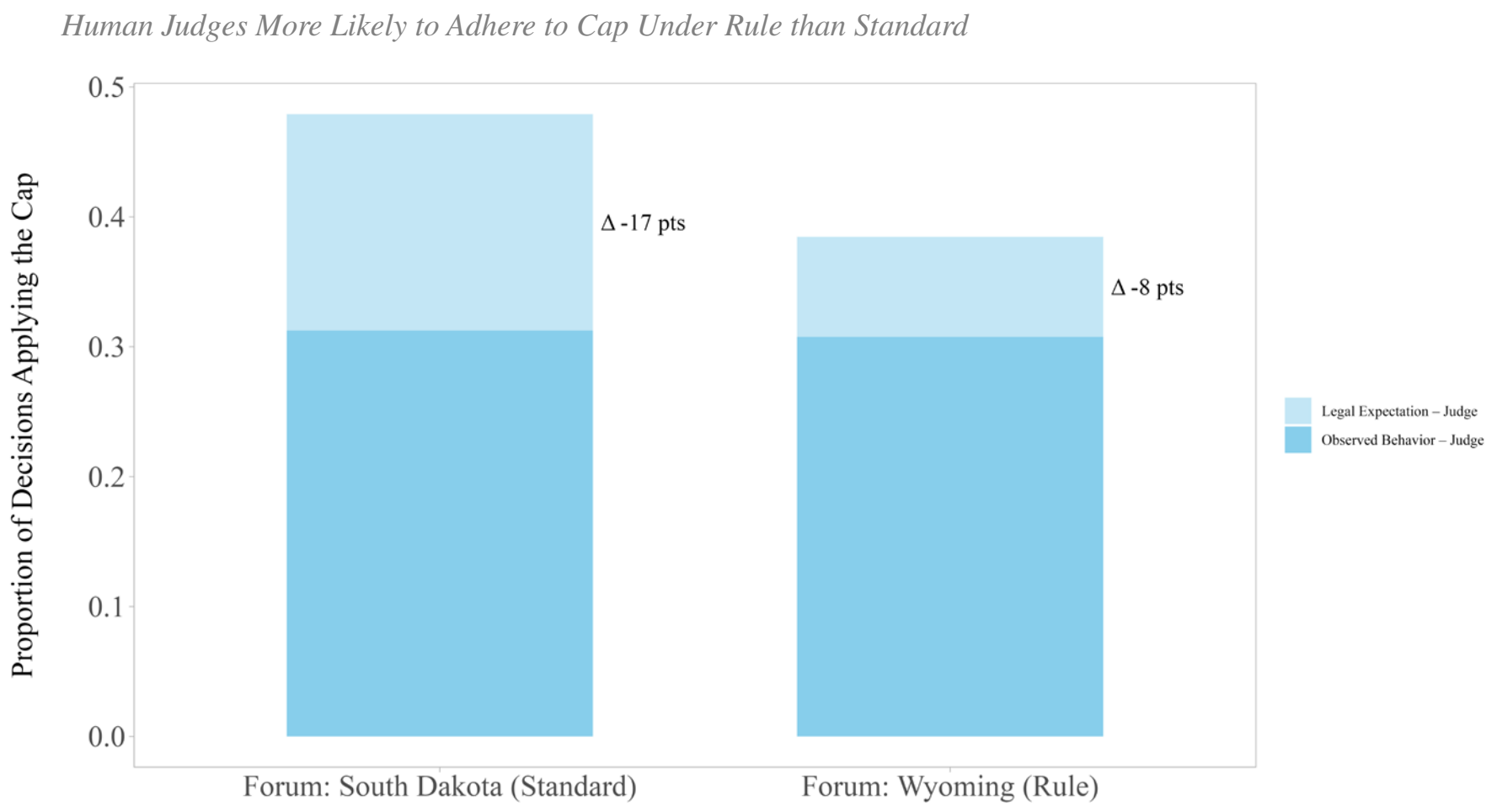

**FIGURE 7** | Proportion of decisions applying the cap vs. legal expectation by forum. Human judges more likely to adhere to cap under rule than standard. [Color figure can be viewed at wileyonlinelibrary.com]

Now consider the South Dakota forum, where judges are required to apply the law of the common domicile:

- If Kansas is the common domicile, then the correct legal decision is to apply Kansas law (cap). Once again, judges face a conflict. Those who want to avoid the cap must violate the standard or (as we argue below) persuade themselves that the standard permits them to deviate from the law of common domicile.
- If Nebraska is the common domicile, the correct legal decision is to apply Nebraska law, once again aligning legal compliance with cap aversion.

In sum, in order for judges to achieve the desirable outcome—avoiding the cap—they must violate the Wyoming rule if the accident occurred in Kansas, but violate the South Dakota standard if it occurred in Nebraska. As noted earlier, our findings and legal theory suggest that standards give judges more discretion than rules. It follows, then, that judges should be more likely to apply the cap when they are subject to rules than when they are subject to standards.

This is indeed what Klerman and Spamann find (Figure 7). When the law directs judges to apply a cap, the gap between the legal expectation and the observed behavior is only 8 percentage points under the rule but jumps to 17 percentage points under the standard. While GPT does not exhibit cap aversion (since it simply always follows the law), we do find that some other models examined in our robustness checks display cap-avoidance behavior similar to judges (discussed below), including an earlier version of GPT (GPT-4o).

## 4.6 | Regression Analysis

To ensure comparability with Klerman and Spamann's results, we replicate their regression design exactly,[16] using the same contrast-coding specifications.[17] The independent variables are the factors tested in the experiment (forum, accident location, and sympathy), as well as their interaction terms. The dependent variables are (1) whether Kansas law was applied (Model 1a), (2) whether the correct legal outcome was reached (Model 1b), and (3) whether the law of the accident location was applied (Model 1c). As Klerman and Spamann emphasize, these three models are algebraically symmetrical. That is, the dependent variables change but the underlying structure and content remain identical, allowing the same results to be interpreted through different lenses (see below). Because our setup involving the judicial data is identical to theirs, the first seven coefficients (as well as the constants) are the exact same as those reported in Klerman and Spamann's Table A2 (p. 120). The only extension is the inclusion of GPT in the analysis, which we introduce via standard dummy-variable coding (GPT = 1, human judge = 0), along with interaction terms that capture how GPT's responses to the experimental variables (Forum, Accident Location, and Sympathy) differ from judges. The full regression results are reported in Table 3, with 95% confidence intervals.

To recap Klerman and Spamann's findings (and to illustrate how these algebraically symmetrical models operate), let us begin by focusing on Model 1b (Following the Correct Law). The forum coefficient is −0.32, which indicates that when moving from the Wyoming forum to the South Dakota forum, judges are 32 percentage points less likely to follow the law. This is consistent with our earlier finding (and with Spamann and Klerman's data) that judges are less constrained by standards than by rules. The same substantive pattern can be interpreted through Models 1a (Applied Kansas Law) and 1c (Applied Lex Loci Delicti). In Model 1a, the accident location coefficient is −0.32, indicating that when the accident location shifts from Kansas to Nebraska, judges are less likely to apply Kansas law. This reflects judges' tendency to show a stronger preference for applying the law of the accident location, a tendency confirmed in Model 1c, where the constant is 0.32, implying that judges apply the law of the accident location 66% of the time ($1/2 * 0.32 + 0.50 = 66\%$), even though it is legally correct to in only 21% of cases.

We now turn our attention to the GPT results, again starting with Model 1b. The GPT coefficient is 0.78, indicating that GPT is 39 percentage points more likely ($0.78/2 = 39\%$) to follow the law than judges are. Given that the model-estimated law-following rate for judges is 61% ($0.22/2 + 0.50 = 61\%$), this implies

**TABLE 3** | Regressions.

| | Model 1a—applied Kansas law | Model 1b—followed correct law | Model 1c—Applied Lex Loci Delicti |
|---|---|---|---|
| Forum (F) | −0.09 [−0.39, 0.21] | −0.32 [−0.62, −0.02] | −0.22 [−0.52, 0.08] |
| SD = 1; WY = −1 | | (*) | |
| Accident Location (AL) | −0.32 [−0.62, −0.02] | −0.09 [−0.39, 0.21] | 0.30 [−0.01, 0.60] |
| NE = 1; KS = −1 | (*) | | |
| Sympathetic party (SP) D = 1; *p* = −1 | 0.21 [−0.09, 0.52] | 0.19 [−0.11, 0.50] | 0.01 [−0.29, 0.32] |
| F × AL | 0.22 [−0.08, 0.52] | −0.30 [−0.60, 0.01] | 0.09 [−0.21, 0.39] |
| F × SP | −0.24 [−0.55, 0.06] | −0.01 [−0.32, 0.29] | −0.19 [−0.50, 0.11] |
| AL × SP | −0.01 [−0.32, 0.29] | −0.24 [−0.55, 0.06] | −0.21 [−0.52, 0.09] |
| F × AL × SP | 0.19 [−0.11, 0.50] | 0.21 [−0.09, 0.52] | 0.24 [−0.06, 0.55] |
| GPT | 0.30 [−0.01, 0.60] | 0.78 [0.48, 1.08] | −0.32 [−0.62, −0.02] |
| | | (***) | (*) |
| GPT: F | 0.09 [−0.21, 0.39] | 0.32 [0.02, 0.62] | −0.78 [−1.08, −0.48] |
| | | (*) | (***) |
| GPT: AL | 0.32 [0.02, 0.62] | 0.09 [−0.21, 0.39] | −0.30 [−0.60, 0.01] |
| | (*) | | |
| GPT: SP | −0.21 [−0.52, 0.09] | −0.19 [−0.50, 0.11] | −0.01 [−0.32, 0.29] |
| GPT: F × AL | 0.78 [0.48, 1.08] | 0.30 [−0.01, 0.60] | −0.09 [−0.39, 0.21] |
| | (***) | | |
| GPT: F × SP | 0.24 [−0.06, 0.55] | 0.01 [−0.29, 0.32] | 0.19 [−0.11, 0.50] |
| GPT: AL × SP | 0.01 [−0.29, 0.32] | 0.24 [−0.06, 0.55] | 0.21 [−0.09, 0.52] |
| GPT: F × AL × SP | −0.19 [−0.50, 0.11] | −0.21 [−0.52, 0.09] | −0.24 [−0.55, 0.06] |
| Constant | −0.30 [−0.60, 0.01] | 0.22 [−0.08, 0.52] | 0.32 [0.02, 0.62] |
| | | | (*) |
| *N* | 261 | 261 | 261 |

*Note:* $^{*}p < 0.05$, $^{**}p < 0.01$, $^{***}p < 0.001$.

a predicted GPT law-following rate of 100%, which matches the results observed in our experiment. Once again, the same result is expressed via Models 1a and 1c. In Model 1c, GPT is 39 percentage points less likely to apply the lex loci delicti rule when the forum switches from Wyoming to South Dakota. This is exactly what one would expect from a perfectly law-following GPT. Under the standard-based doctrine, GPT no longer applies the rule of the accident location. In Model 1a, GPT is 39 percentage points more likely than judges to apply Kansas law when the accident location shifts from Kansas to Nebraska and the forum shifts from Wyoming to South Dakota (once again, expected from a perfectly law-following GPT).

Another important result concerns the GPT x Forum interaction in Model 1b. The coefficient here is 0.32, indicating that switching from judges to GPT increases the forum effect by 0.32 on the −1 (did not follow the law) to +1 (followed the law) scale. Judges, however, exhibit a negative forum effect of −0.32, meaning that GPT's true forum effect would be 0. Given GPT's invariant legal compliance, this is exactly what we would expect. GPT's likelihood of following the law does not vary based on the forum because it follows the law in every instance. This pattern generalizes across all the experimental factors. GPT's main effects for forum, accident location, and sympathy (as well as all their interaction terms) are all equal in magnitude and opposite in sign to the corresponding judicial effects. (This pattern is true in both Models 1a and 1c, as well.) The regressions thus represent our earlier findings regarding the LLM's extreme formalism.[18]

## 5 | Reasoning

### 5.1 | Quality Analysis

Both the judges and GPT labored under certain disadvantages when they wrote the “opinions” that accompanied their decision. The judges were given a limited amount of time to decide the case and write their opinion. GPT was limited by its lack of human faculties. Still, most of the judges and GPT (in all of its decisions) wrote reasonable opinions. We provide two examples below.

The scenario in both cases involved the SD-KS-D condition: South Dakota forum, Kansas accident location (and thus Nebraska common domicile), and a sympathetic defendant. In these examples, the human judge and GPT both correctly ruled that Nebraska law applied. The judge wrote as follows:

> The district court, sitting in South Dakota, is required to apply the choice of law rules adopted by the South Dakota state courts. The South Dakota Supreme Court, following the majority of other courts, adopted the Restatement (Second) most significant relationship approach (rejecting the place of injury approach) in resolving multi-state torts conflicts of law. In Burhenn, a case mentioned only in passing by the defendant, the South Dakota Supreme Court focused on the center of the relationship between the parties, in applying the most significant relationship test in a wrongful death case. Restatement Section 145, Comment (d) suggests that the local law of the parties' common domicile should control if that was both the state from which the trip started and the state to which the parties intended to return. Here, Nebraska was the state where the trip began and where the trip was intended to end. Nebraska was also, at the time of the crash, the state of domicile of both parties. Restatement Section 145, Comment (e), Illustration 1 further supports the conclusion that, applying South Dakota's choice of law principles, the factors mentioned lead to the conclusion that the center of the relationship between these parties was Nebraska.[19]

GPT-5 wrote:

> Under Klaxon, South Dakota's choice-of-law rules govern. South Dakota follows the Restatement (Second) and applies the law of the state with the most significant relationship to the particular issue. For damages—a loss-distribution issue—the Restatement §171 and comment b favor the parties' common domicile, especially when their relationship is centered there. At the time of the accident both Parker and Rogers were Nebraska domiciliaries, and their relationship was centered in Nebraska; Rogers's post-accident move to South Dakota does not alter that analysis. Although the conduct and injury occurred in Kansas, those contacts alone do not control the damages issue. South Dakota precedent (Chambers v. Dakotah Charter; Selle v. Pierce) applies the common-domicile state's law in analogous loss-distribution conflicts, and the Peters decision cited by defendant is distinguishable and not controlling. Nebraska therefore has the greater interest in the measure of damages, and Nebraska law applies.[20]

As is typical in these cases, neither the judge nor GPT mentions the personal characteristics of the plaintiff or defendant that were used to portray one as more sympathetic than the other. As in most cases, they do not mention the impact of the cap on limiting the plaintiff's recovery.

### 5.2 | Error Analysis With GPT-4o

We also manually reviewed the opinions. We were curious whether GPT provided incoherent or erroneous explanations for its decisions even though those decisions were correct. After all, the role of the judge is not only to make a decision but to explain it to the litigants in an opinion that provides guidance for the others. In fact, GPT's opinions were clear, accurate, and well-written in all cases. In an earlier version of our experiment, however, we used GPT-4o, which did commit errors in some decisions and in the opinions. Because these results may reveal something about how LLMs commit errors and how they have evolved, we report them in Appendix A.1.

## 6 | Robustness and Methodological Concerns

There are several important concerns when using LLMs in experimental research. Some scholars argue that the causal inference logic in traditional human-subject experiments doesn't transfer over to LLM-based designs. Gui and Toubia (2025), for instance, highlight that the assumption that only the manipulated variables change, while all else remains constant, may not hold true in LLM experiments. Altering a single variable might have unintended consequences in the model's internal representation of the scenario because, as prediction engines rather than cognitive agents, LLMs tend to "fill in" missing context based on patterns in their training data. As a result, small prompt changes may shift implicit assumptions in ways that introduce confounds, reducing ecological validity and making causal claims less reliable.

To test for potential confounding, we follow Gui and Toubia's two key recommendations: (1) explicitly informing the LLM that it is part of an experiment, and (2) using unambiguously causal prompts (e.g., instructing the LLM to directly consider sympathy). The first approach aims to prevent the model from inferring its own assumptions about context; the second increases the likelihood that the model's output reflects statistically grounded reasoning tied directly to the variable being manipulated, rather than just surface-level "blinded" probabilistic associations. We implemented both adjustments in a separate replication of our experiment[21] to test robustness and obtained identical results, with the model continuing to follow the law 100% of the time. This also addresses concerns raised by other scholars regarding the sensitivity of LLMs to subtle prompt changes (Blair-Stanek and Van Durme 2025; Choi 2027; Waldon et al. 2025).

Another concern is that the use of repeated runs does not reflect genuine sample variability (Choi, forthcoming). Because the model's architecture remains consistent across runs, each output represents a statistical perturbation rather than a truly distinct perspective. (In this manner, it is more akin to sampling different responses from the same individual than drawing from a diverse population.) We agree with this critique. However, our objective is not to simulate 200 “individual” judges. Instead, we seek to measure the “average” response of the model across varying conditions. This approach allows us to detect not only overall outcome differences but also patterns in legal reasoning across completions. By comparing these results to the benchmark results from Klerman and Spamann's original experiment, this setup allows us to assess (1) whether LLMs are capable of consistently producing plausible legal arguments, and (2) whether those arguments markedly differ from those of human judges.

Nonetheless, in an effort to increase sample variability, we replicate the experiment using various other LLMs, including Gemini 3 Pro, Gemini 2.5 Pro, o4-mini, Llama 4 Maverick, GPT-4o, Llama 4 Scout and GPT-4.1. Like GPT-5, the Gemini models, o4-mini, and Llama 4 Maverick are “reasoning” models, which are designed to outperform standard models in tasks requiring logical structure and multi-step logic (Bergmann 2025). Llama 4 Scout, GPT-4o, and GPT-4.1, on the other hand, are non-reasoning models. We find that our results are, by and large, robust across the reasoning models, but not across the non-reasoning ones. Table 4 presents our full robustness results.

Gemini 3 Pro, like GPT-5, followed the law in every single instance. That is, across all conditions and iterations, Gemini 3 reached the same exact outcomes as GPT-5, exhibiting no forum, accident location, or sympathy effects. Gemini 2.5 Pro and GPT-4o also demonstrated strong legal compliance, following the law in 92% and 90% of cases, respectively. While o4-mini and Llama 4 Maverick didn't quite match the performances of GPT-5 and the Gemini models, both models still outperformed judges, adhering to the law in 79% and 75% of cases, respectively. Llama 4 Scout and GPT-4.1 are the only models that underperformed judges in following the law, and no model exhibited a sympathy effect.

Further analysis reveals a couple of interesting patterns. First, several models (Gemini 2.5 Pro and Llama 4 Maverick) were more likely to follow the law if it took the form of a rule (Wyoming) rather than a standard (South Dakota). This discrepancy is similar to what we observed with judges and GPT-4o and aligns with the generally accepted view that standards offer judges more flexibility than do rules. Second, some models (Llama 4 Maverick and Llama 4 Scout) exhibited cap-avoidance behavior similar to that observed among judges and GPT-4o. For the LLMs that exhibited a comparable cap-aversion, it is possible that the models' training data reflect a broader view that damage caps, to some degree, conflict with the pursuit of justice and should be disregarded when possible.

Overall, our robustness checks reveal a clear hierarchy. The most advanced models (GPT-5 and Gemini 3 Pro) follow the law religiously, while the remaining reasoning models and GPT-4o markedly outperform human judges. Significant deviation from the law is largely confined to the less capable models, and even there, noncompliance mirrors patterns observed in human judges, such as greater adherence to rules than standards and aversion to damage caps.

Finally, one might wonder why we chose to replicate judicial experiments rather than evaluate LLMs using real cases. There are several reasons. First, judicial experiments allow for the systematic manipulation of both legally relevant and irrelevant variables (sympathy, doctrine, etc.), something that is impossible to control in real cases. Given how little is understood about LLMs and their internal decision-making processes, the primary objective of this paper is exploratory: identifying the specific factors that LLMs respond (and don't respond) to. Second, experiments provide a benchmark for evaluating legal compliance, because the “legally correct” outcome across conditions is predetermined by the experimental design. In real cases, it is more ambiguous what the “legally correct” outcome is, let alone whether the judge actually reached it. Third, real cases introduce contamination risk, since the judicial opinions may have appeared in the models' training data. While there are methodological choices that can be taken to mitigate this risk, they typically involve tradeoffs (e.g., relying on older models). Lastly, real cases are subject to selection effects: most cases are resolved without ever having an opinion published, making published cases an imperfect sample of the broader universe of litigation. While evaluation of real cases is nonetheless valuable, it falls outside the scope of this paper. In ongoing work, we explore this avenue (Posner and Saran Forthcoming).

**TABLE 4** | Results for different models.

| Model | Reasoning model | Forum effect | Accident location effect | Sympathy effect | % of Decisions following the law | Who follows law more |
|---|---|---|---|---|---|---|
| GPT-5 (original) | Yes | No | No | No | 100% | LLM |
| Gemini 3 Pro | Yes | No | No | No | 100% | LLM |
| Gemini 2.5 Pro | Yes | Yes | No | No | 92% | LLM |
| o4-mini | Yes | No | Yes | No | 79% | LLM |
| Llama 4 Maverick | Yes | Yes | Yes | No | 75% | LLM |
| GPT-4o | No | Yes | Yes | No | 90% | LLM |
| Llama 4 Scout | No | No | Yes | No | 50% | Judges |
| GPT-4.1 | No | Yes[a] | No | No | 50% | Judges |

[a]GPT-4.1 exhibits a forum effect in the opposite direction to the forum effect observed in other models, as well as the general behavior of human judges. That is, GPT-4.1 was actually more constrained by the standard than it was by the rule; in fact, the model followed the standard 100% of the time and followed the rule 0% of the time. One possible explanation is a “common domicile bias,” where the model believes the law of the parties' common domicile is the correct legal outcome, regardless of what the governing doctrine is.

## 7 | Discussion

In our earlier *Judge AI* paper, we found that GPT was more formalistic than judges in the sense that GPT consistently follows the relevant precedent and was not influenced by nonlegal factors—in that setting, the extent to which the defendant was portrayed sympathetically or unsympathetically. GPT's decisions were similar to those of students who were subjects in a second experiment. This pattern raised several questions. Should we treat the human judges as the benchmark for evaluating LLMs or should the benchmark be formal compliance with the law? That students comply with the law more than professional judges raises the possibility that socially desirable judging requires judges to deviate from formal rules to avoid injustices or bad policy outcomes, a staple of legal reality theory.

After our prompt-engineering efforts failed to nudge GPT to decide cases like the professional judges rather than students, we speculated that it may be impossible for LLMs to be designed to act like human judges. First, LLMs are trained on texts, including judicial opinions, and judges do not always (usually?) provide their real reasons for decisions in opinions, while many texts, including textbooks, for example, provide idealized descriptions of the rule of law. Under this idealized description, judges follow the law; they do not make it. Second, LLMs are subject to guardrails imposed by their designers, including what appears to be a rule-of-law guardrail, and it is not clear that guardrails could be established that enable AI judges to take into account nonlegal factors in the way that human judges apparently do. Third, we live in a pluralistic society in which disagreement is tolerated rather than suppressed. Judges have different experiences, ideologies, and moral commitments, and it is hard to imagine that this kind of variety could be replicated by AIs. Nor does it seem likely that political elites or ordinary voters would be willing to tolerate being judged by a machine over which they cannot exercise influence through appointments and other mechanisms, or fight for the opportunity to do so.

In this paper, similar questions arise. While we have so far followed Klerman and Spamann in describing deviations from the formal law as "errors," the truth may be more complicated.

Let us return to Klerman and Spamann's experiment. The materials[22] provided to the subjects in the South Dakota conditions describe the Restatement test. According to that test, the law of the state with 'the most significant relationship to the occurrence and the parties' is used in tort disputes.[23] The Restatement continues:

> Contacts to be taken into account … include (a) the place where the injury occurred, (b) the place where conduct causing the injury occurred, (c) the domicile [and] residence … of the parties, and (d) the place where the relationship, if any, between the parties is centered.[24]

The materials distributed to the experimental subjects also include two South Dakota Supreme Court cases that adopt a comment in the Restatement that says:

> When the plaintiff and defendant are domiciled in the same state, and particularly if in addition there is a special relationship between them which is centered in this state, it would seem that this state is likely to be the state with the most significant relationship with respect to the issue of damages.[25]

Those two cases—Chambers v. Dakotah Charter[26] and Selle v. Pierce[27]—both applied the law of the state of common domicile. In the materials provided to the subjects, parties who seek to apply the law of common domicile describe the cases as follows:

> Chambers v. Dakotah Charter, 488 N.W.2d 63, 67 (S.D. 1992) is the case in which the South Dakota Supreme Court adopted the Restatement (Second)'s most significant relationship approach to choice of law. The suit involved a South Dakota plaintiff who sued a South Dakota corporation for injuries the plaintiff sustained in Missouri while falling from the steps of defendant's bus. Although Missouri was the place of injury, the Supreme Court of South Dakota applied South Dakota's law of comparative negligence. The court emphasized that "South Dakota was the domicile, residence, place of incorporation, and place of business of the parties, as well as the place where the relationship of the parties was centered." Id. at 68.

> In Selle v. Pierce, 494 N.W.2d 634 (S.D. 1993), a Nebraska resident filed a defamation action against another Nebraska resident regarding a defamatory letter mailed to South Dakota. Even though the plaintiff's business and social reputation in South Dakota was injured by the defamation, the Court applied Nebraska law, which barred punitive damages, because the parties resided in Nebraska and the relationship between plaintiff and defendant was "centered" in Nebraska. Id. at 637.

The opposing party argues that the cases are distinguishable—roughly on the ground that in those cases the tort or the injury is spread across several states, whereas in the instant case, the tort and injuries were concentrated in a single state—Kansas or Nebraska. The opposing party also cites Peters.[28] The Hawaii court applied Hawaii law rather than the law of common domicile because (among other things) the parties bought insurance from a Hawaii rental car agency and the insurer would have expected Hawaii law to apply.

Our view is that the Restatement standard, as embodied in South Dakota law, is not, as Klerman and Spamann argue, a common domicile "rule" that permits only one outcome.[29] The Restatement itself, the cases, and commentary discussed in the materials permit deviation from common domicile for a variety of reasons—even including the forum court's view of which state has the "best law." We are thus not sure that a South Dakota court that applies accident location is necessarily wrong. Consider an excerpt from one such (human) opinion:

> Parties, although domiciled at the time of the accident in Nebraska, did not share a special relationship such as husband and wife, + defendant subsequently changed domicile. His time of residence in Nebraska may have been fluid + accidental, on account of his incarceration.
>
> Plaintiff was hospitalized in Kansas for an extended period, + received rehabilitative treatment in Kansas.
>
> Rental insurance contract which provided $1 million coverage, anticipated that Kansas law would be applied, although this is not evident from the evidence presented in any event since the balance of contacts seems to weigh in favor of Kansas, it seems apparent that since the car was rented in Kansas, the inherited property is in Kansas, + the accident was in Kansas, that Kansas law should apply. All medical + rehabilitative treatment was received in Kansas, + presumably the Kansas facilities would seek reimbursement from the tortfeasor through subrogation.
>
> The shared domicile argument seems trivial in view of the number of Kansas contacts, + is insubstantial as a basis for application of Nebraska law to the circumstances.[30]

If the South Dakota standard permits judges to deviate from common domicile, then it is hard to conclude that human judges in the Nebraska forum erred when they applied Kansas law. They simply exercised judgment differently from the other judges. And, if that is correct, the gap between judges and the LLMs in applying the law may be smaller than it first appears.

If the proper interpretation of the results is that human judges deviated from the South Dakota law, we propose two explanations. First, the subjects committed a cognitive error—most commonly, misidentifying the state in which the plaintiff resides at the time of the accident. Second, the subjects deliberately deviated from the law because they believed that it would be unjust to deprive the plaintiff of a full recovery for serious injuries. The South Dakota standard provided "cover" for doing justice, or maybe it required the judges to do justice.

Such cap aversion was not demonstrated in our original replication with GPT-5, though it did appear with other LLMs during our robustness checks as well as our earlier experiment with GPT-4o. The similarity between these LLMs and judges is important because judges deviating from a formalistic view of the law, for reasons such as a perceived unfairness of capped damages, are a classic example of legal realism; they step out of the pure legal rule-following sphere to think about what impact their judgement will have and rule accordingly. For the LLMs that exhibited a comparable tendency to avoid caps, it is possible that their training data may reflect this broader view that damage caps, to some degree, do conflict with the pursuit of justice. On this interpretation, an AI judge may be willing to depart from the law but only when the law is vague enough to camouflage that departure.

Whatever may explain such behavior in judges and some LLMs, however, certainly does not apply to GPT-5 and Gemini 3 Pro. Across all conditions, regardless of doctrinal flexibility, both models followed the law without fail. To the extent that LLMs are evolving over time, the direction is clear: error-free allegiance to formalism rather than the humans' sometimes-bumbling discretion that smooths away the sharper edges of the law. And does that mean that LLMs are becoming better than human judges or worse?

**Acknowledgments**

We thank participants at the AI in Social Science conference (University of Chicago), American Law and Economics Association Annual Meeting, 6th Machine Lawyering Conference (Chinese University of Hong Kong) and Conference on Empirical Legal Studies. Thanks also to Holger Spamann, Joshua Fischman, Christoph Engel, and Valeria Hans, and to Bennett Bunten for helpful research assistance. Data necessary to replicate the results of this article are available upon request from the corresponding author.

**Endnotes**

[1] When we began this project, we used the then-most advance OpenAI model, GPT-4o. GPT-4o did make some errors, but fewer than judges at a statistically significant level. GPT-4o's performance is discussed in the Appendix A.

[2] These included circuit, district, magistrate, and bankruptcy judges. Demographic details of judges can be found at Table 1, Klerman and Spamann (2024).

[3] In 2019, the Kansas Supreme Court struck down the state's damage cap as unconstitutional, ruling that it violated a plaintiff's right to a jury trial since determination of damages is a core function of the jury, and a cap undermines that authority. Notably, Klerman and Spamann's experiment was conducted in 2017 and 2018 (before the ruling) and their material reflected Kansas law as it stood at the time (i.e., with the cap in place). In our paper, when we refer to Kansas' cap, we are referring to the pre-2019 cap as presented in the original experiment materials. Given that our experiment was conducted in 2025, one might wonder whether Kansas' updated law is reflected in GPT's training data and thus skews its decisions. However, we find no evidence of such contamination. Across multiple iterations, we provided the LLM with the full experimental materials and then asked whether Kansas imposes a cap on non-economic damages. Each time, the model responded that Kansas has a cap, consistent with how Kansas law is presented in the experimental materials. We also performed a continuation test (see, for example, Golchin and

Surdeanu 2024), providing the model with the opening excerpt of the Kansas Supreme Court decision invalidating the damage cap, and then asking the model to complete the passage. The model consistently failed to do so. As a final check, we reformatted the test as a fill-in-the-blank (FITB) task, where the model, once again, consistently failed to complete the passage. A related concern is that the model may have encountered Klerman and Spamann's (2024) paper during training and, as a result, already "knew" what the legally correct outcomes were. To assess this possibility, we again provided the LLM with the complete experimental materials and then asked it to identify the academic paper or project from which they were drawn. Across multiple iterations, the model was unable to connect the materials to Klerman and Spamann's work, nor was it able to accurately identify their source (on multiple occasions, it confidently, yet incorrectly, attributed them to Cass Sunstein).

[4] We return to this aspect of the experiment design infra.

[5] In the hypothetical, the defendant moves to either South Dakota or Wyoming after the accident, which is why the suit is brought in one of those states, despite the accident location and common domicile (at the time of accident) being elsewhere.

[6] For a complete description of experimental manipulation related to party sympathy, see Table 3, Klerman and Spamann (2024).

[7] As with any judicial experiment, Klerman and Spamann's (2024) design simplifies the adjudication process. Judges deciding real cases have access to clerks, opportunities to deliberate with colleagues, and substantially more time than what was provided in the experiment. Accordingly, their performance in the experiment may not fully capture judicial decision-making in practice. Our comparison between LLMs and judges should be considered within this context.

[8] See Posner and Saran (2026) for a discussion of the temperature parameter and system prompts.

[9] The complete user and system prompt are reported in Appendix A.1. The full set of prompts across all experimental conditions and robustness tests will be available at an online Appendix A.

[10] Klerman & Spamann, Online Appendix A.

[11] See *Best v. Taylor Machine Works* (1997); *Klutschkowski v. Oregon Medical Group* (2013).

[12] See *Morris v. Savoy* (1991); *Lucas v. United States* (1988).

[13] See *Ferdon v. Wisconsin Patients Compensation Fund* (2005); *Beason v.* I.E. *Miller Services Inc*. (2019).

[14] Klerman and Spamann do not report statistically significant differences in judge behavior across these conditions, for reasons including the study's small sample size. Accordingly, we cannot definitively conclude that a bias against damage caps exists. Still, this framing offers a plausible explanation for why legal compliance varies depending on where the accident occurred.

[15] For a detailed comparison of cap compliance and aversion by Forum and Accident Location, see Table A1. For a breakdown across all conditions (Forum, Accident Location, and Sympathy), see Table A2. These tables are exact replications of Tables 4 and A1 from Klerman and Spamann (2024), but with our LLM results presented alongside theirs for human judges.

[16] Independent variables are contrast coded as such: Forum (South Dakota: 1, Wyoming: –1), Accident Location (Nebraska: 1, Kansas: –1), and Sympathy (Defendant: 1, Plaintiff: –1). The dependent variables are contrast coded as such: Model 1a (Application of Kansas Law (cap): 1, Application of Nebraska Law (no cap): –1), Model 1b (Legally Correct Outcome: 1, Legally Incorrect Outcome: –1), Model 1c (Application of Accident Location Law: 1, Application of Common Domicile Law: –1).

[17] Contrast coding is particularly useful in factorial experiments because it averages effects between levels of a factor, rather than requiring a (sometimes arbitrary) baseline from which to measure deviations. Coefficients, as a result, directly represent the effect of switching levels. See Wiens and Nilsson (2016).

[18] Further formal tests (Fisher's exact) are reported in Table A3, which also affirm the stark difference between judges' behavior and GPT's formalism.

[19] ID #242.

[20] ID #26.

[21] We implemented the adjustments in the model's system prompt: "You are a federal judge participating in a research experiment involving judicial decision-making. You will be provided with a hypothetical automobile accident case. In this scenario, (1) the forum of adjudication, (2) the accident location (and common domicile), and (3) the sympathetic characteristics of the plaintiff and defendant are deliberately specified and should be treated as given. Your task is to determine whether Kansas or Nebraska law applies to the calculation of damages and to provide a response in JSON format."

[22] See Klerman & Spamann, Online Appendix, OA.10–11.

[23] Restatement (Second) of Conflict of Laws § 145(1).

[24] Id at § 145(2).

[25] Restatement (Second) Conflict of Laws § 171 cmt. b.

[26] 488 N.W.2d 63 (S.D. 1992).

[27] 494 N.W.2d 634 (S.D. 1993).

[28] Peters v. Peters.

[29] After discussing the law in South Dakota, they argue:

> That is, one might say that in South Dakota's application of the Restatement 2nd's standard, "common domicile," has morphed into a rule…, even while the Restatement 2nd's open-ended language offers more wiggle room for a judge inclined not to follow it.

Klerman and Spamann (2024), 114.

[30] ID #216.

[31] Peters v. Peters, 63 Haw. 653, 654, 634 P.2d 586, 587 (1981).

[32] For human judges, it is hard to make comparisons because there were many more South Dakota than Kansas cases.

## Appendix A

### Fraction Choosing Kansas Law (Cap) by Forum and Accident State

| | | Factor 2 | | Total by Factor 1 |
|---|---|---|---|---|
| | | Accident location/common domicile | | |
| | | Accident in NE/common domicile in KS | Accident in KS/common domicile in NE | |
| Factor 1 Forum (place of adjudication) | WY (place-of-accident rule) | Legal implication: 0% | Legal implication: 100% | |
| | | **GPT:** 0% (0/50) | **GPT:** 100% (50/50) | **GPT:** 50% (50/100) |
| | | **Judges:** 13% (1/8) | **Judges:** 60% (3/5) | **Judges:** 31% (4/13) |
| | SD (common domicile standard) | Legal implication: 100% | Legal implication: 0% | |
| | | **GPT:** 100% (50/50) | **GPT:** 0% (0/50) | **GPT:** 50% (50/100) |
| | | **Judges:** 26% (6/23) | **Judges:** 36% (9/25) | **Judges:** 31% (15/48) |
| Total by Factor 2 | | **GPT:** 50% (50/100) | **GPT:** 50% (50/100) | **GPT:** 50% (100/200) |
| | | **Judges:** 23% (7/31) | **Judges:** 40% (12/30) | **Judges:** 31% (19/61) |

*Note:* This table replicates tab. 4 of Klerman and Spamann (2024) with GPT results added.

### Fraction Choosing Kansas Law (Cap) by Sympathy, Forum, and Accident State

| | | Factor 3 | | | |
|---|---|---|---|---|---|
| | | Sympathy | | | |
| | | Sympathetic defendant and unsympathetic plaintiff | | Sympathetic plaintiff and unsympathetic defendant | |
| Factor 2 accident/domicile | | Accident NE/domicile KS | Accident KS/domicile NE | Accident NE/domicile KS | Accident KS/domicile NE |
| Factor 1 Forum | WY (rule) | **GPT:** 0% (0/25) | **GPT:** 100% (25/25) | **GPT:** 0% (0/25) | **GPT:** 100% (25/25) |
| | | **Judges:** 25% (1/4) | **Judges:** 100% (2/2) | **Judges:** 0% (0/4) | **Judges:** 33% (1/3) |
| | SD (standard) | **GPT:** 100% (25/25) | **GPT:** 0% (0/25) | **GPT:** 100% (25/25) | **GPT:** 0% (0/25) |
| | | **Judges:** 33% (4/12) | **Judges:** 25% (3/12) | **Judges:** 18% (2/11) | **Judges:** 46% (6/13) |
| Total by Factor 3 | | **GPT:** 50% (50/100) | | **GPT:** 50% (50/100) | |
| | | **Judges:** 33% (10/30) | | **Judges:** 29% (9/31) | |

*Note:* This table replicates tab. 5 of Klerman and Spamann (2024) with GPT results added.

### Randomization-Based Fisher Tests

| | | Main effect by forum (Factor 1) | | Difference |
|---|---|---|---|---|
| | | WY (rule) | SD (standard) | |
| | | ($N$=100) | ($N$=100) | |
| Testing effect of | Law (accident location) (Factor 2) | **GPT:** $\Delta=1.00$ ($p=0.00$) | **GPT:** $\Delta=-1.00$ ($p=0.00$) | **GPT:** $\Delta$ (WY)—$\Delta$ (SD)$=2.00$ ($p=0.00$) |
| | | **Judges:** $\Delta=0.48$ ($p=0.14$) | **Judges:** $\Delta=-0.09$ ($p=0.76$) | **Judges:** $\Delta$ (WY)—$\Delta$ (SD)$=0.56$ ($p=0.06$) |
| | Sympathy (Factor 3) | **GPT:** $\Delta=0.00$ ($p=1.00$) | **GPT:** $\Delta=0.00$ ($p=1.00$) | **GPT:** $\Delta$ (WY)—$\Delta$ (SD)$=-0.00$ ($p=1.00$) |
| | | **Judges:** $\Delta=0.36$ ($p=0.17$) | **Judges:** $\Delta=-0.04$ ($p=1.00$) | **Judges:** $\Delta$ (WY)—$\Delta$ (SD)$=0.40$ ($p=0.12$) |

*Note:* This table replicates tab. 7 of Klerman and Spamann (2024) with GPT results added.

**GPT-5 Simulate Judicial Decision Call**

*System Prompt:* "You are a federal district judge reviewing an automobile accident case. Your task is to determine whether Kansas or Nebraska law applies to the calculation of damages and provide a response in JSON format."

*User Prompt:* Please read the following instructions:

{Condition}

Now that you have read Judge Simmons' findings and the parties' memoranda, please review the texts of the cited cases, statutes, and other materials

{Cited_Material}

Now that you have read all of the relevant information, please provide a decision on whether Kansas or Nebraska law applies to the calculation of damages (indicate this by either saying "Kansas" or "Nebraska"), as well as a brief 1-paragraph description of your rationale. Respond in JSON format with two keys: Decision and Rationale.

**GPT-4o Error Analysis**

We limited ourselves to cases in which judges reached incorrect outcomes. (In the correctly decided cases, the reasoning in the opinions was mostly correct or at least largely sensible—indicating that judges put at least some effort in trying to decide the cases correctly.) We start with the decisions in the South Dakota forum conditions—where the applicable doctrine is a standard rather than a rule—as these are the only settings in which GPT-4o deviates from the law. The tables below document the types and frequencies of errors (or statements that may indicate deviation from the law) in these cases.

**TABLE A1** | South dakota forum: Subject chooses Nebraska; common domicile is Kansas (GPT-4o vs. Judges).

| AI decisions | Fraction | Percentage | Example |
|---|---|---|---|
| Factual error | 2/21 | 9.52% | "Unlike the cases of common domicile, this case does not involve both parties having the same domicile at the time of the accident, as the defendant is now domiciled in South Dakota." ID #74 [Both parties were domiciled in the same state at the time of the accident.] |
| Common domicile error | 12/21 | 57.14% | "While the defendant argues for the application of Kansas law due to common domicile before the accident, the current domiciles are split between Kansas and South Dakota." ID #51 [Common domicile is determined at time of accident, not time of trial.] |
| Mentions cap | 7/21 | 33.33% | "Furthermore, Nebraska's lack of a cap on non-economic damages aligns with the notion of compensating the victim where the injury primarily occurred, consistent with sections 145 and 171 of the Restatement (Second)." ID #58 |
| Peters v. Peters mentioned | 6/21 | 28.57% | "Additionally, having procured insurance in Nebraska's jurisdiction suggests a reliance on its laws, paralleling precedent set in Peters v. Peters." ID#61 |
| Judge decisions | Fraction | Percentage | Example |
| Factual error | 1/17 | 5.88% | "The restatement 2 has additional factors but on balance most of them lean toward applying Nebraska law. For instance, our defendant no longer resides in South Dakota where the plaintiff resides." ID #226 [The plaintiff has never resided in South Dakota.] |
| Common domicile error | 9/17 | 52.94% | "The third contact factor does not help because the parties are domiciled in two different states-Kansas and South Dakota." ID#221 [Common domicile is determined at the time of the accident, not time of the trial.] |
| mentions cap | 0/17 | 0.00% | N/A |
| Peters v. Peters Mentioned | 0/17 | 0.00% | N/A |

**Common Domicile Confusion With Timing**

The most frequent type of error involved a misapplication of the "common domicile" factor, which under South Dakota law and the Restatement (Second) § 145 and § 171 should be determined based on the parties' domicile at the time of the accident. In Table A1, 57% of GPT-4o's incorrect decisions (12 of 21) and 53% of judges' incorrect decisions (9 of 17) stated that the parties were domiciled in different states, based on post-accident relocations. These responses fail to recognize that, for purposes of the Restatement, domicile is assessed at the time of the tortious event. This kind of error suggests not factual sloppiness, but a misunderstanding of the doctrine's temporal focus.

### Common Domicile Confusion with Facts

A smaller subset of decisions involved straightforward factual mistakes—misstating a party's residence, misidentifying the location of the accident, or misunderstanding elements of the record. GPT-4o made factual errors in 2 out of 21 incorrect decisions (9.5%), while judges did so in 1 of 17 (5.9%). These errors may reflect either noise in GPT-4o's internal representation of the facts or misreading of the scenario, and they resemble the kinds of inattentive mistakes that real-world judges occasionally make.

### Cap-Driven Reasoning

A third (7 of 21, or 33%) of GPT-4o's incorrect rulings explicitly referred to the absence of a damages cap in Nebraska as part of the rationale for applying Nebraska law—even though Kansas law should have applied under a proper reading of the standard. Human judges made no such references in their incorrect decisions. The presence of this factor in GPT's reasoning suggests a possible normative bias: an implicit preference for outcomes that avoid applying capped damages. "Cap aversion" is consistent with the empirical pattern across both humans and AI, and may not represent a doctrinal mistake per se, but a normative position that an open-ended standard allows the judge to take into account the "best law," as the Restatement acknowledges.

### Citation of (or Reliance on) Inapt Precedent

About a quarter (6 of 21, or 29%) of GPT-4o's incorrect decisions mentioned *Peters v. Peters*, a Hawaii case that applied the law of accident location to a tort dispute where a couple from New York were involved in an accident in Hawaii. Although the Hawaii court discussed the Restatement standard, it held that the "preferred choice-of-law analysis is an assessment of the interests and policy factors involved with a purpose of arriving at a desirable result in each situation," an eye-wateringly vague standard.[31] None of the incorrect human decisions cited *Peters*, which may suggest that GPT-4o relies excessively on superficially relevant precedents when seeking justification.

### Other Scenarios

We also examined the reasoning of judges in settings where GPT-4o avoided errors. Tables A2–A4 below summarize what we found. Again, the common domicile issue tripped up judges. In South Dakota, a judge should choose Nebraska law if the common domicile is Nebraska, but 56% of the judges who erroneously selected Kansas argued that the parties did not share a domicile in Nebraska. In Wyoming, the rule-based choice-of-law doctrine resulted in few errors. In the two instances where judges chose Nebraska law when they should have chosen Kansas law, both opinions invoked a "public policy exception" to the lex loci rule to justify avoiding Kansas's damage cap. Here, cap aversion—not mere confusion—may explain the result.

**TABLE A2** | South Dakota Forum—Kansas accident, chooses Kansas.

| Judge decisions | Fraction | Percentage | Example |
|---|---|---|---|
| Factual error | 1/9 | 11.11 | "His (the defendant) time of residence in Nebraska may have been fluid + accidental, on account of his incarceration." ID# 212 [The case materials state that both parties have lived in Nebraska their whole life, not because of incarceration.] |
| Common domicile error | 5/9 | 55.56 | "(3) the domicile of the parties does not require an outcome either way, because Ms. Parker lives in Nebraska and Mr. Rogers lives in South Dakota." ID #201 [Common domicile is determined by location at time of accident, not location at time of trial.] |
| Mentions cap | 1/9 | 11.11 | "As a non-legal, but still a possible thought on the sympathy scale (I know, I know [sic]) the difference in the total award between the states is not enormous." ID #210 |
| Peters v. Peters mentioned | 1/9 | 11.11 | "As for cases, the most comparable case is Peters v. Peters, in which similar policy considerations existed for the state in which the conduct and injury occurred." ID #244 |

**TABLE A3** | Wyoming Forum–Nebraska accident location, chooses Kansas.

| Judge decisions | Fraction | Percentage | Example |
|---|---|---|---|
| Factual error | 0/1 | 0 | N/A |
| Mentions cap | 0/1 | 0 | N/A |

**TABLE A4** | Wyoming Forum–Kansas accident location, chooses Nebraska.

| Judge decisions | Fraction | Percentage | Example |
|---|---|---|---|
| Factual error | 0/2 | 0 | N/A |
| Mentions cap | 2/2 | 100 | "However, on the facts of this case, the public policy exception to the lex loci delicti rule must be applied on the issue of noneconomic damages." ID# 254 |

These tables underscore a possible interpretation of the results: when legal errors do occur—especially for GPT-4o—they are disproportionately concentrated in standard-based conditions, not rule-based ones.[32] Moreover, the nature of the errors reinforces the interpretive ambiguity of standards: they allow plausible but (apparently) legally incorrect reasoning to masquerade as doctrinal fidelity. In some cases, especially where the standard appears to permit justice-oriented reasoning (e.g., avoiding caps), GPT-4o's and judges' deviations may reflect principled judgments rather than errors. But in others—particularly those involving temporal misapplication of domicile—there is little doubt that the reasoning is simply flawed.

Our results for GPT-5, however, suggest that GPT-4o was less powerful than GPT-5, which is of course what one would expect. To the extent that we can generalize to LLMs, our findings suggest that more advanced models will make fewer errors and/or display less of the human propensity to deviate from the rules in order to achieve social goals.